\documentclass[fleqn,10pt]{wlscirep}
\usepackage[utf8]{inputenc}
\usepackage[T1]{fontenc}
\usepackage{xcolor}
\usepackage{colortbl}
\usepackage[noend]{algpseudocode}
\usepackage{algorithmicx,algorithm}
\usepackage{multirow}
\usepackage{booktabs}
\usepackage{float} 
\usepackage{booktabs}
\usepackage{tabularx}
\usepackage{geometry}
\usepackage{hyperref}
\title{CORA: Generalizable coronary artery disease assessment and risk stratification from coronary CT angiography using pathology-centric representation learning}

\author[1]{Jinkui Hao}
\author[1]{Gorkem Durak}
\author[1]{Halil Ertugrul Aktas}
\author[1,2]{Ulas Bagci}
\author[1]{Bradley D. Allen}
\author[3,4]{Nilay S. Shah}
\author[1,*]{Bo Zhou}

\affil[1]{Department of Radiology, Northwestern University, Chicago, IL, USA}
\affil[2]{Department of Biomedical Engineering, Northwestern University, Evanston, IL, USA}
\affil[3]{Department of Cardiology, Northwestern University, Chicago, IL, USA}
\affil[4]{Department of Preventive Medicine, Northwestern University, Chicago, IL, USA}

\affil[*]{\textit{bo.zhou@northwestern.edu}}

\begin{abstract}
Coronary artery disease, a leading cause of cardiovascular mortality worldwide, can be assessed non-invasively by coronary computed tomography angiography (CCTA). Although deep learning has advanced automated CCTA analysis, clinical translation remains constrained by the scarcity of expert-annotated data and by the spatial sparsity of coronary pathology, which occupies only a small fraction of each scan. Widely used label-free pretraining strategies, such as masked image modeling and contrastive learning, optimize for global anatomical reconstruction and tend to under-represent these tiny localized pathological features. Here we present CORA (\textbf{CO}ronary \textbf{R}epresentation learning via \textbf{A}bnormality synthesis), an annotation-efficient model for comprehensive coronary artery disease assessment. Rather than reconstructing background anatomy, CORA learns from volumetric CCTA through a synthesis-driven self-supervised strategy: an anatomy-guided engine inserts diverse synthetic calcified and non-calcified lesions into unlabeled scans, reframing pretraining as an abnormality-detection task that biases representation learning toward clinically relevant disease features. We pretrained CORA on 10,138 unlabeled CCTA volumes and evaluated it across datasets from nine independent hospitals. Across plaque characterization, stenosis detection, and coronary artery segmentation, CORA consistently outperformed strong self-supervised pretraining baselines, with the largest gains on external multi-center data, indicating robust generalization under distributional shift. Coupling the imaging encoder with structured clinical variables further enabled near-term major adverse cardiac event (MACE) risk stratification. Our results show that pathology-centric, synthesis-driven pretraining is an effective and scalable strategy for annotation-efficient coronary artery disease assessment from CCTA.

\end{abstract}
\begin{document}

\flushbottom
\maketitle
\thispagestyle{empty}

Coronary artery disease (CAD), driven by atherosclerotic plaque accumulation within the epicardial vasculature, is among the leading causes of cardiovascular mortality worldwide and accounts for the largest share of ischemic heart disease deaths each year~\cite{lindstrom2022global}. The clinical use of coronary computed tomography angiography (CCTA) for non-invasive evaluation of symptomatic patients is rapidly growing, and its utility for primary atherosclerotic cardiovascular disease prevention is an active area of research~\cite{writing20212021,narula2024prospective,jukema2025diagnostic}. The unique value of CCTA lies in its ability to visualize not only luminal stenosis but also the underlying atherosclerotic plaque morphology, a critical and independent predictor of future myocardial infarction~\cite{williams2020low,feuchtner2025ai}. Landmark randomized trials, including SCOT-HEART and PROMISE, have shown that CCTA-guided management reduces major adverse cardiac events and improves long-term clinical outcomes~\cite{scot2018coronary,douglas2015outcomes}.
Despite this established efficacy, the clinical translation of CCTA remains constrained by a substantial interpretation burden: expert analysis of a single examination can require on the order of 25 minutes of radiologist time~\cite{lin2022deep}, and the growing demand for cardiac imaging continues to strain reader capacity~\cite{thomsen2016characteristics,nurmohamed2024development}. This bottleneck can delay patient access to diagnostic imaging and contributes to inter-observer variability, particularly for high-risk plaque features that are spatially subtle and easily overlooked during routine review~\cite{pinna2025machine}.
 
A central obstacle to automated CCTA analysis is the extreme spatial sparsity of the pathological signal. Atherosclerotic plaques typically occupy only a small fraction of the total CCTA volume~\cite{shrivastava2025systematic}, embedded within a complex background of myocardium, extra-cardiac fat, and contrast-enhanced blood pool. This imbalance fundamentally challenges standard representation learning: models trained on such data are intrinsically biased toward the dominant anatomical structures that constitute the vast majority of image content, while systematically under-representing the subtle, spatially localized intensity variations that characterize clinically decisive vascular abnormalities. Consequently, current approaches often rely on multi-stage pipelines involving vessel detection, centerline extraction, and multi-planar reformatting~\cite{maurovich2014comprehensive, ma2021transformer, lee2020differences,denzinger2019coronary,lorenzatti2024interaction}, which limits scalability and introduces error propagation across stages.

The high cost of expert annotation has motivated strong interest in self-supervised learning (SSL), which derives transferable representations from large collections of unlabeled data~\cite{gui2024survey,zhang2024challenges,awais2025foundation}. In 2D medical imaging, SSL-pretrained models have shown impressive downstream capabilities~\cite{zhou2023foundation, chen2024towards, wang2024pathology, ma2024pretraining, tejada2025nicheformer, yao2025eva}. However, extending these paradigms to 3D volumetric CCTA exposes a conceptual limitation. Dominant 3D SSL objectives~\cite{zhou2021models, chen2023masked,yan2022sam, wald2025revisiting, wang2025sam}, such as masked image modeling, emphasize global anatomical reconstruction, and emerging evidence suggests they offer limited benefit for downstream cardiovascular tasks because they sacrifice the representation of sparse, localized pathological signals in favor of background anatomy~\cite{wald2025openmind}.

\begin{figure*}[t]
    \centering{  
    \includegraphics[width=1\linewidth]{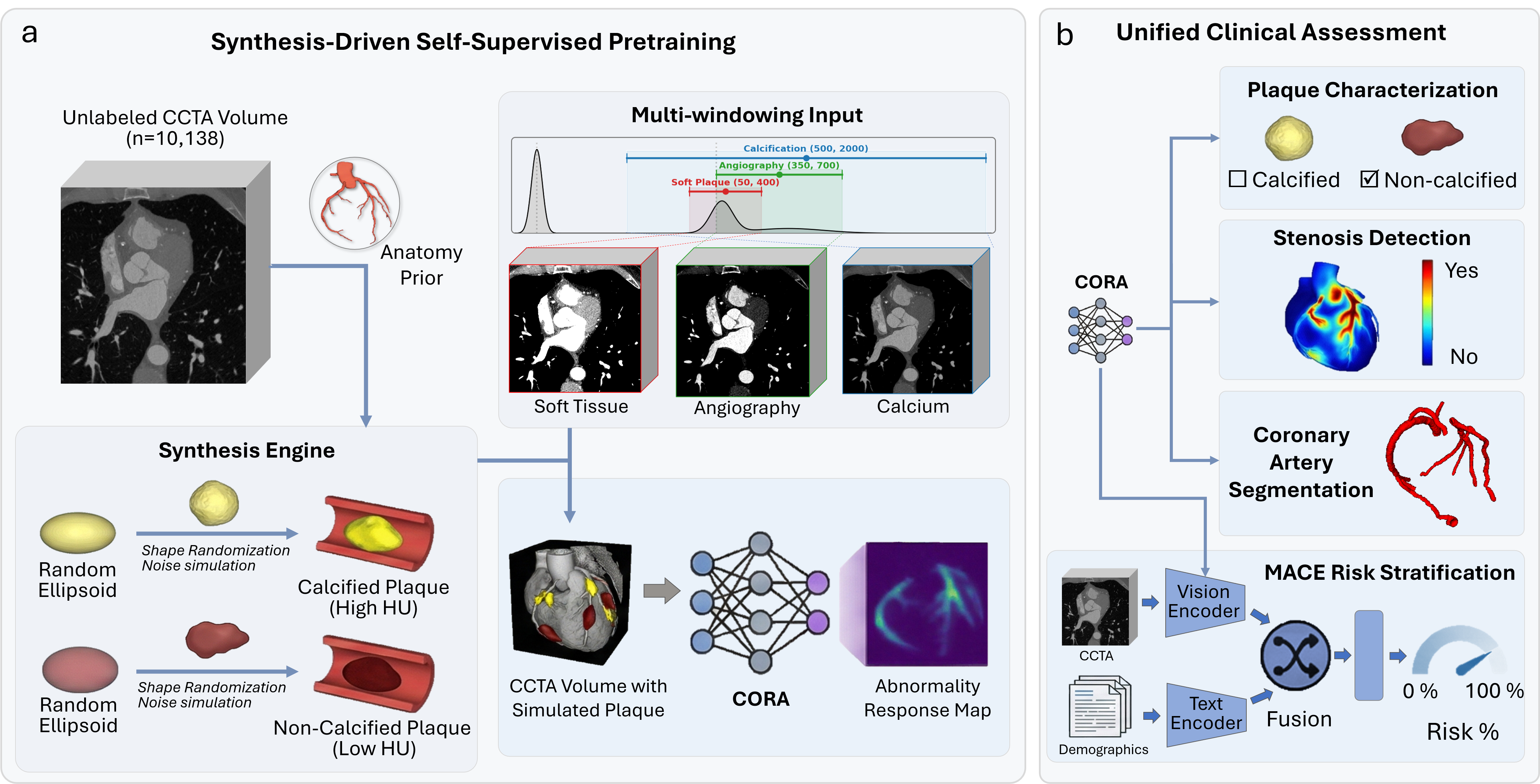}
    }
    \caption{ Overview of CORA and synthesis-driven pretraining framework. (a) Synthesis-driven self-supervised pretraining of CORA on large-scale unlabeled CCTA volumes. An anatomy-guided lesion synthesis engine generates diverse calcified and non-calcified plaque patterns with controlled morphology and attenuation, which are inserted into CCTA volumes to create simulated abnormality. Multi-windowing inputs capture complementary tissue characteristics across soft tissue, contrast-enhanced lumen, and calcium windows, enabling the encoder to learn pathology-centric representations and produce localized abnormality response maps. (b) Down-stream clinical assessment using the pretrained CORA encoder. A single foundation model supports multiple downstream tasks, including coronary plaque classification, stenosis detection, and coronary artery segmentation. For prognostic modeling, imaging representations are further integrated with patient demographic and clinical information through a multimodal fusion framework to enable major adverse cardiac event (MACE) risk stratification. 
}
    \label{fig-1}
\end{figure*}

Here we develop CORA, an annotation-efficient model for comprehensive CAD assessment that addresses these limitations through two complementary innovations. First, in place of generic anatomical reconstruction, CORA adopts a synthesis-driven self-supervised strategy: an anatomy-guided engine generates realistic calcified and non-calcified plaque morphologies with controlled Hounsfield Unit distributions and inserts them into unlabeled CCTA volumes, explicitly reframing pretraining as an abnormality-detection task. This biases the learned representations toward clinically relevant vascular pathology rather than dominant background anatomy. Second, CORA replaces single-channel normalization with a multi-windowing input that jointly encodes soft-tissue, contrast-enhanced lumen, and calcium characteristics, enabling discrimination between low-attenuation non-calcified plaques and high-density calcified lesions within a single forward pass, without vessel segmentation or multi-planar reformatting.

We pretrained CORA on a large-scale cohort of 10,138 unlabeled CCTA volumes and systematically evaluated it across multi-center datasets from nine hospitals. We benchmarked CORA on diagnostic tasks (plaque characterization and stenosis detection) and an anatomical task (coronary artery segmentation), comparing against training from scratch and strong 3D self-supervised pretraining baselines. We then examined whether the learned imaging representations, combined with structured clinical variables, support near-term MACE risk stratification. Across all evaluated tasks, CORA outperformed competing pretraining strategies, with the largest margins observed on external multi-center cohorts. Our results show how pathology-centric, synthesis-driven pretraining provides a scalable and annotation-efficient strategy for unified coronary artery disease assessment from CCTA.

\section*{Results}

\subsection*{Synthesis-driven pretraining biases CORA toward clinically relevant vascular pathology}
To evaluate CORA across clinically diverse settings, we benchmarked it against baseline training from scratch and three state-of-the-art 3D vision foundation models, including MAE~\cite{he2022masked}, VolumeFusion~\cite{wang2023mis}, and VoCo~\cite{wu2025large}, across four task categories: volume-level coronary plaque characterization, lesion-level stenosis detection, dense 3D coronary artery segmentation, and 30-day major adverse cardiac event (MACE) risk stratification. Evaluations spanned multi-center datasets from nine independent hospitals, enabling assessment of cross-site generalizability under realistic distributional shift (Fig.~\ref{fig-1}).

Conventional self-supervised learning strategies, including masked image modeling and contrastive learning, optimize for global anatomical reconstruction, implicitly prioritizing the dominant background structures that constitute the vast majority of CCTA image content. In contrast, CORA employs a synthesis-driven self-supervised paradigm: an anatomy-guided lesion synthesis engine generates realistic calcified and non-calcified plaque morphologies and inserts them into the 12,801 unlabeled CCTA pretraining volumes, explicitly biasing representation learning toward the spatially sparse vascular pathology that governs downstream clinical tasks. CORA further employs a multi-windowing input strategy that simultaneously encodes soft-tissue, contrast-enhanced lumen, and calcium window characteristics, enabling discrimination of both plaque subtypes within a single forward pass.

\begin{table}[ht!]
\centering
\caption{Baseline demographic, clinical, and imaging characteristics of the training and validation cohorts used for coronary plaque identification and MACE risk stratification.}
\label{tab-1}
\small
\begin{tabular}{lccc}
\toprule
\textbf{Characteristic} & \textbf{Training Set} & \textbf{Internal Test} & \textbf{External Test} \\
 & ($N=2,220$) & ($N=555$) & ($N=2,108$) \\
\midrule
\textbf{Demographics} & & & \\
Age (years), mean $\pm$ SD & $63.1 \pm 13.8$ & $62.0 \pm 14.4$ & $63.7 \pm 13.5$ \\
Female sex, $n$ (\%) & $935 (42.1\%)$ & $215 (38.7\%)$ & $1,025 (48.6\%)$ \\
Race, $n$ (\%) & & & \\
\quad White & $1,438 (64.8\%)$ & $338 (60.9\%)$ & $1,723 (81.7\%)$ \\
\quad Black or African American & $349 (15.7\%)$ & $103 (18.6\%)$ & $126 (6.0\%)$ \\
\quad Asian\textsuperscript{a} & $98 (4.4\%)$ & $26 (4.7\%)$ & $82 (3.9\%)$ \\
\quad Other or Unknown\textsuperscript{b} & $335 (15.1\%)$ & $88 (15.8\%)$ & $177 (8.4\%)$ \\
\midrule
\textbf{Clinical Characteristics} & & & \\
BMI ($kg/m^2$), mean $\pm$ SD & $29.0 \pm 7.7$ & $29.2 \pm 7.3$ & $29.8 \pm 7.2$ \\
Systolic BP (mmHg), mean $\pm$ SD & $127.8 \pm 19.3$ & $127.1 \pm 18.9$ & $128.6 \pm 19.7$ \\
Diastolic BP (mmHg), mean $\pm$ SD & $72.4 \pm 12.8$ & $72.3 \pm 12.6$ & $73.3 \pm 12.2$ \\
Tobacco Use, $n$ (\%) & & & \\
\quad Never & $1,041 (46.9\%)$ & $255 (45.9\%)$ & $1,031 (48.9\%)$ \\
\quad Former & $430 (19.4\%)$ & $113 (20.4\%)$ & $560 (26.6\%)$ \\
\quad Current & $114 (5.1\%)$ & $27 (4.9\%)$ & $136 (6.5\%)$ \\
Hypertension within 1 year, $n$ (\%) & $990 (44.6\%)$ & $264 (47.6\%)$ & $1,062 (50.4\%)$ \\
Diabetes mellitus, $n$ (\%) & $318 (14.3\%)$ & $81 (14.6\%)$ & $339 (16.1\%)$ \\
\midrule
\textbf{Laboratory Metrics} & & & \\
Total Cholesterol (mg/dL) & $170.0 \pm 47.0$ & $171.8 \pm 47.1$ & $171.0 \pm 73.5$ \\
LDL Cholesterol (mg/dL) & $96.5 \pm 43.8$ & $98.3 \pm 37.9$ & $93.7 \pm 37.4$ \\
HDL Cholesterol (mg/dL) & $49.7 \pm 17.3$ & $50.3 \pm 17.5$ & $51.9 \pm 17.1$ \\
\midrule
\textbf{Imaging Findings} & & & \\
Non-calcified Plaque present, $n$ (\%) & $963 (43.4\%)$ & $237 (42.7\%)$ & $935 (44.4\%)$ \\
Calcified Plaque present, $n$ (\%) & $1,300 (58.6\%)$ & $326 (58.7\%)$ & $1,390 (65.9\%)$ \\
\midrule
\textbf{Clinical Outcome} & & & \\
30-day MACE, $n$ (\%) & $192 (8.6\%)$ & $48 (8.6\%)$ & $237 (11.2\%)$ \\
\bottomrule
\addlinespace
\multicolumn{4}{l}{\textsuperscript{a}\footnotesize Includes Asian Indian, Chinese, Filipino, Korean, Vietnamese, Japanese, and Other Asian.} \\
\multicolumn{4}{l}{\textsuperscript{b}\footnotesize Includes American Indian, Pacific Islander, Declined, and Unknown.} \\
\end{tabular}
\end{table}

\subsection*{CORA achieves superior plaque characterization with robust cross-site generalization}

Accurate characterization of coronary plaque phenotype is central to cardiovascular risk stratification, as plaque composition predicts the likelihood of future rupture and acute coronary syndrome.
We fine-tuned and evaluated CORA on this task as a volume-level multi-label classification problem predicting the simultaneous presence of non-calcified and calcified plaques.
Training and internal test cohorts were derived from Northwestern Memorial Hospital (N=2,220 and 555, respectively); the external test cohort comprised 2,108 patients from eight geographically and demographically distinct hospitals.
The dataset characteristics used for fine-tuning are summarized in Table~\ref{tab-1}. 

\begin{figure*}[!t]
    \centering{  
    \includegraphics[width=1\linewidth]{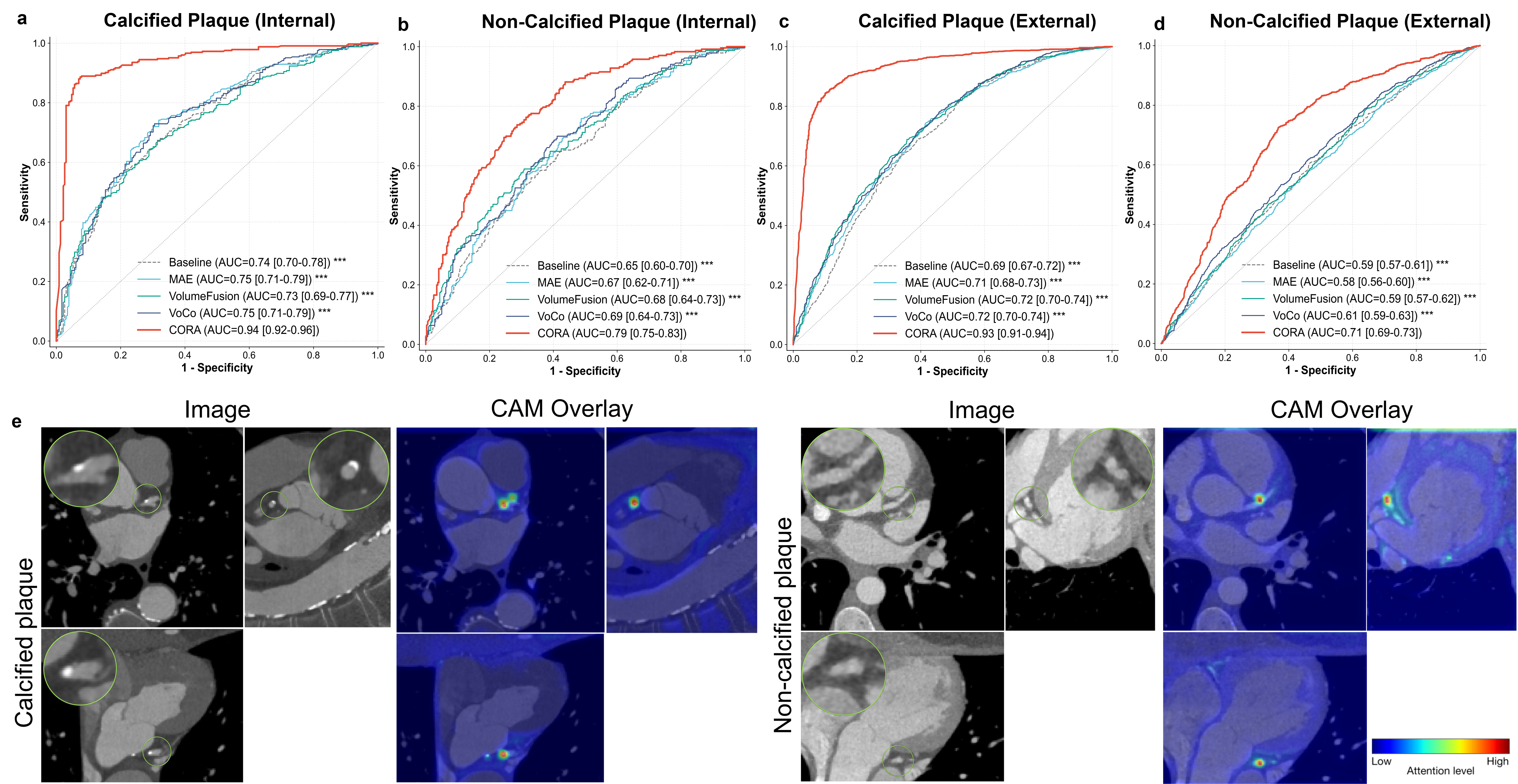}
    }
    \caption{ Performance of plaque characterization across internal and external cohorts. Receiver operating characteristic (ROC) curves for the detection of calcified and non-calcified coronary plaques on the internal test cohort (a,b) and external multi-center cohort (c,d). CORA consistently outperformed models trained from scratch and existing 3D foundation models, with the largest performance gains observed on the external cohort, indicating superior generalization under domain shift. e, Visualization of model attention using Grad-CAM showed model accurately identified plaque regions in CCTA volumes.
    The 95\% CIs for AUCs were computed using 500 bootstrap resamples. Differences in AUCs were assessed using a bootstrap test. Statistical significance is denoted as $*p < 0.05$, $**p < 0.01$, and $***p < 0.001$.
}
    \label{fig-2}
\end{figure*}

Across both internal and external test sets, CORA consistently outperformed all comparison methods (Fig.~\ref{fig-2}). On the internal cohort, CORA achieved AUROCs of 0.94 (95\% CI: 0.92–0.96) for calcified plaque and  0.79 (95\% CI: 0.75–0.83) for non-calcified plaque, substantially exceeding the baseline trained from scratch (0.74 [0.70–0.78] and 0.65 [0.60–0.70], respectively; both p$<$ 0.001). CORA demonstrated consistent performance advantages over MAE, VolumeFusion, and VoCo across both plaque subtypes (p$<$ 0.001), reflecting its capacity to encode the distinct attenuation signatures of each phenotype.

Performance advantages were further amplified on the external test cohort, where CORA achieved AUROCs of 0.93 (95\% CI: 0.91–0.94) for calcified plaque and 0.71 (95\% CI: 0.69–0.73) for non-calcified plaque. By contrast, competing methods consistently fell below 0.70 AUROC on external evaluation: MAE achieved 0.71 (95\% CI: 0.68–0.73) and 0.58 (95\% CI: 0.56–0.60), VolumeFusion 0.72 (95\% CI: 0.70–0.74) and 0.59 (95\% CI: 0.57–0.62), and VoCo 0.72 (95\% CI: 0.70–0.74) and 0.61 (95\% CI: 0.59–0.63) for non-calcified and calcified plaque, respectively (all comparisons vs. CORA p $<$ 0.0001). These findings demonstrate that pathology-centric synthesis-driven pretraining enables CORA to learn transferable representations that are robust to site-specific imaging variability.

To investigate the interpretability of our model, we employed Grad-CAM~\cite{selvaraju2017grad} to visualize the diagnostic signals driving the model’s predictions. As shown in Fig.~\ref{fig-2}e, the attention maps reveal that CORA consistently directs its focus to the coronary artery regions, exhibiting the highest activation levels precisely at the site of suspected plaques. For non-calcified lesions, the model's attention aligns with low-attenuation areas along the vessel wall, even in cases with poor contrast-to-noise ratios. For calcified plaques, the activations are localized to high-intensity voxels. This targeted attention demonstrates that the model has internalized a physiologically grounded understanding of coronary anatomy and pathology, rather than relying on spurious background correlations or extra-cardiac noise.

\subsection*{Pathology-aware pretraining sensitizes CORA to subtle coronary stenosis}
\begin{figure*}[!t]
    \centering{  
    \includegraphics[width=1\linewidth]{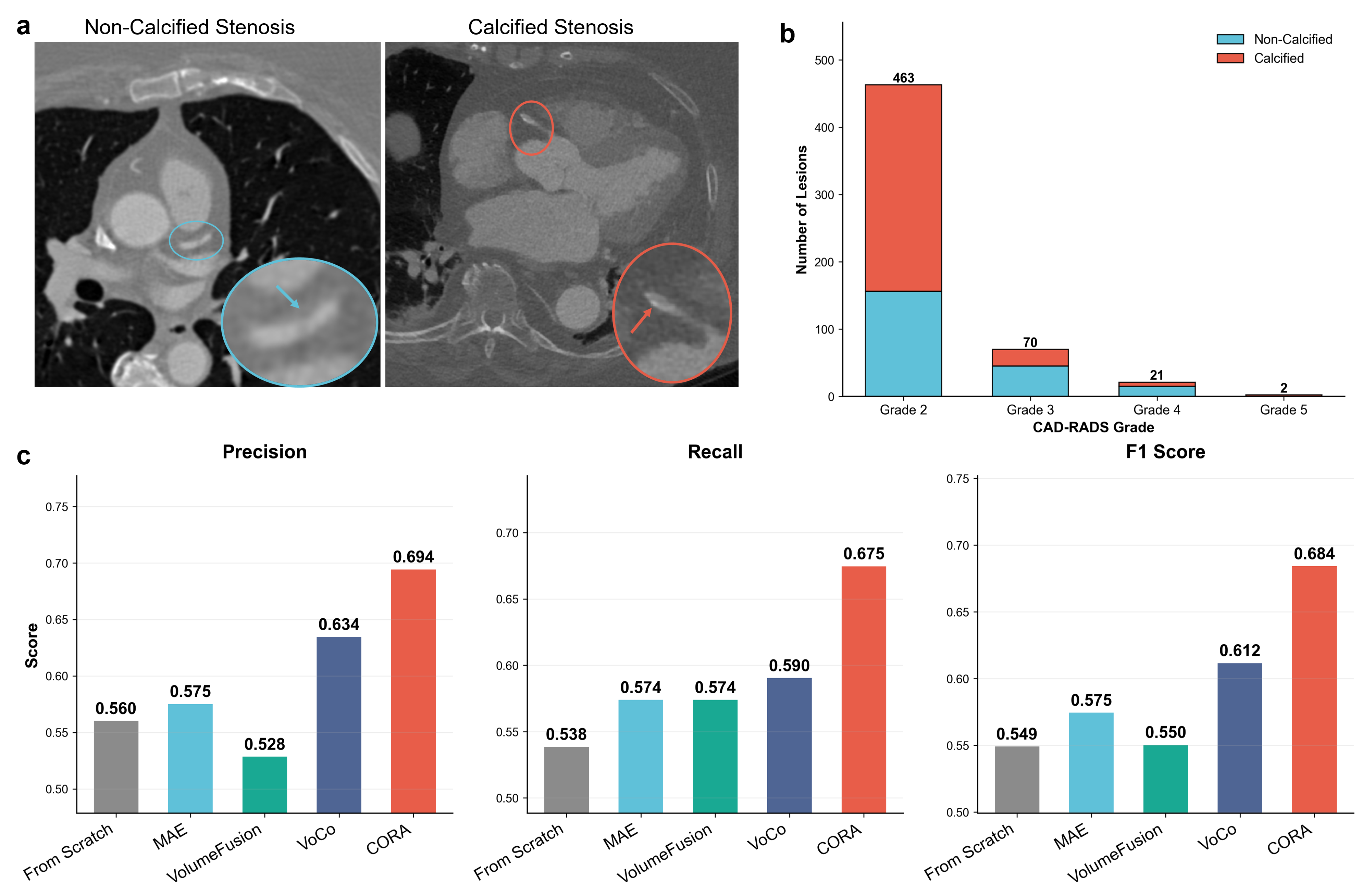}
    }
    \caption{CORA performance in coronary stenosis detection. (a) Representative axial CCTA images illustrating the target pathologies. Magnified insets detail a low-contrast non-calcified stenosis (left, blue) and a calcified stenosis (right, red). (b) Distribution of annotated stenoses in the entire fine-tuning dataset, stratified by CAD-RADS severity grade and plaque composition (calcified, red; non-calcified, blue).   (c) Quantitative benchmarking against a baseline (From Scratch) and self-supervised 3D foundation models (MAE, VolumeFusion, VoCo). CORA (red) achieves state-of-the-art performance across Precision, Recall, and F1 Score.
}
    \label{fig-3}
\end{figure*}

Automated lesion-level stenosis detection is essential for CAD-RADS grading and guides decisions on downstream invasive coronary angiography, yet it remains among the most challenging CCTA analysis tasks due to the morphological heterogeneity of calcified versus non-calcified plaques, the limited spatial resolution of standard CCTA acquisitions, and the absence of large curated annotated datasets. Fig.~\ref{fig-3}a illustrates the distinct morphological features of low-contrast non-calcified plaques and high-attenuation calcified lesions that often complicate automated detection.

We evaluated stenosis detection at the lesion level on 348 CCTA volumes comprising 556 annotated stenotic lesions (340 calcified, 216 non-calcified) across CAD-RADS grades 2–5 (Fig. ~\ref{fig-3}b). A predicted lesion was scored as correct if it spatially overlapped with a ground-truth annotation. CORA attained an overall Precision of 0.694, Recall of 0.675, and F1-score of 0.684, compared with 0.560, 0.538, and 0.549 for the training from scratch baseline. Among 3D foundation models, CORA exceeded MAE (F1=0.575; +19\% relative), VolumeFusion (F1=0.550; +24\%), and VoCo (F1=0.612; +12\%) (Fig.~\ref{fig-3}c).


\subsection*{CORA improves coronary artery segmentation efficiency across data regimes}

Dense 3D coronary artery segmentation provides the anatomical scaffold required for stenosis grading, fractional flow reserve estimation, and surgical planning. It remains technically demanding due to the fine vessel calibre, complex branching architecture, and low contrast of distal coronary segments.
We evaluated CORA on the ImageCAS benchmark~\cite{zeng2023imagecas}, which contains expert-annotated coronary artery masks for approximately 1,000 CCTA volumes. We reserved 40 cases for validation and 60 for testing.

\begin{figure*}[!t]
    \centering{  
    \includegraphics[width=1\linewidth]{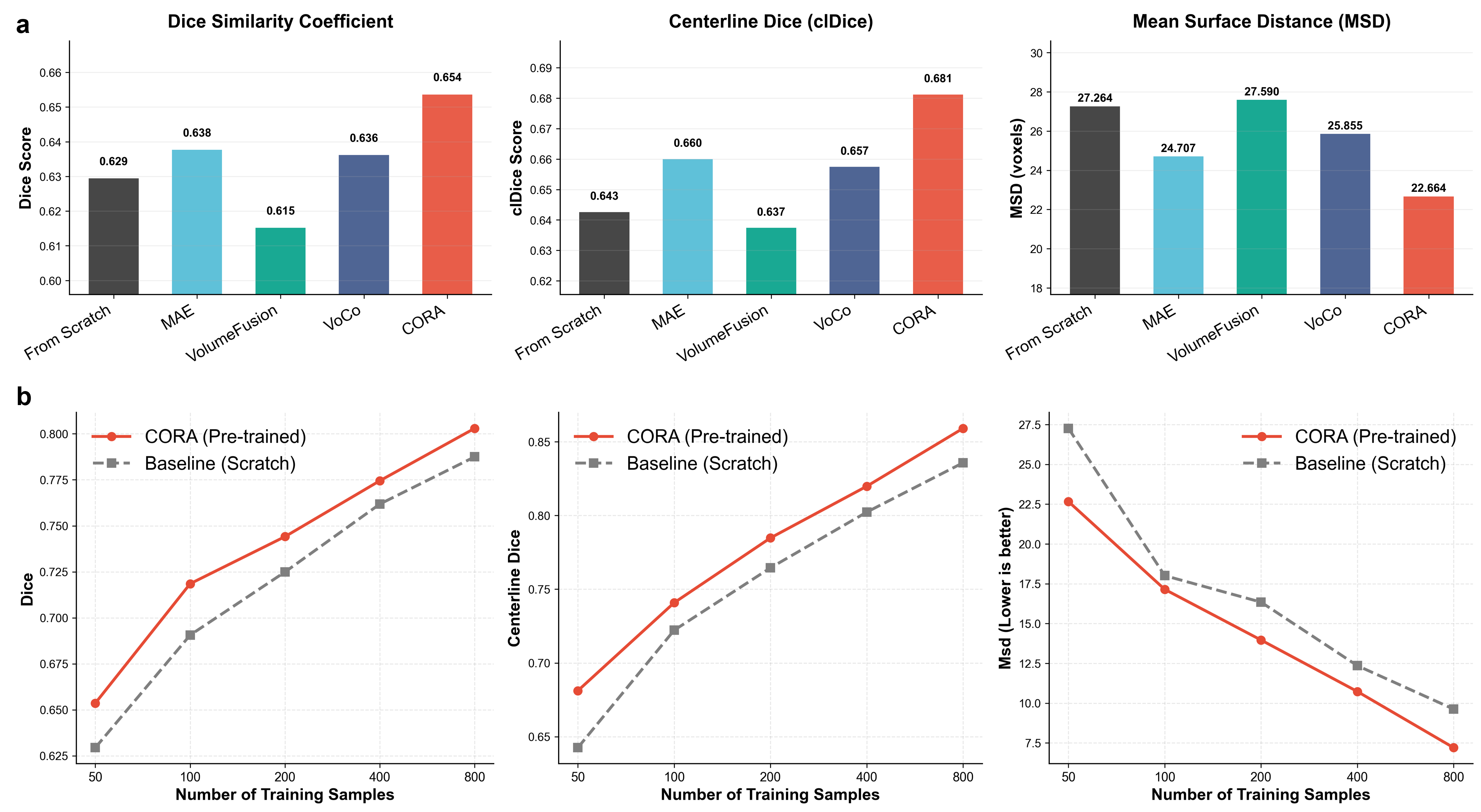}
    }
    \caption{ Coronary artery segmentation performance and data efficiency. (a) Segmentation performance on the ImageCAS test set using 100 labeled training samples, evaluated by Dice similarity coefficient, centerline Dice (clDice), and mean surface distance (MSD). CORA outperformed models trained from scratch and existing 3D foundation models across all metrics. (b) Data efficiency analysis across varying numbers of labeled training samples (50–800). CORA consistently achieved higher segmentation accuracy and better centerline preservation than the baseline, with the largest gains observed in low-data regimes.
}
    \label{fig-4}
\end{figure*}

In the primary evaluation using 100 labeled training samples, CORA achieved a Dice Similarity Coefficient of 0.654 , Centerline Dice (clDice) of 0.681, and Mean Surface Distance (MSD) of 22.664 voxels, outperforming the baseline training from scratch across all metrics (Dice=0.629, clDice=0.651, MSD=27.264) and established 3D foundation models including MAE, VolumeFusion, and VoCo, particularly in vessel continuity as captured by clDice (Fig.~\ref{fig-4}a). The 17\% reduction in MSD (22.664 vs. 27.264 voxels) reflects more accurate delineation of vessel boundaries, which is consequential for downstream quantitative lumen analysis.

We further examined data efficiency by varying the number of labeled training samples (50, 100, 200, 400, and 800). As shown in Fig.~\ref{fig-4}b, CORA consistently outperformed the baseline across all training set sizes, with particularly pronounced gains in the low-data regime. 
Notably, although CORA's pretraining paradigm specifically targets coronary plaque abnormalities, the pretraining significantly improved anatomical segmentation performance. This cross-task benefit suggests that pathology-aware pretraining objectives inherently refine the model's geometric understanding of vascular morphology.

\subsection*{Imaging representations from CORA improve near-term cardiac risk stratification}
For patients presenting with acute chest pain, the 30-day window represents a high-risk period for major adverse cardiac events (MACE), defined as cardiac death, myocardial infarction, or revascularization~\cite{schlett2011prognostic}. We evaluated whether the imaging representations learned by CORA carry prognostic value for near-term MACE, using the cohort described in Table 1 (training N=2,220; internal test N=555; external test N=2,108; 30-day MACE rates 8.6\%, 8.6\%, and 11.2\%, respectively).

\begin{figure*}[!t]
    \centering{  
    \includegraphics[width=1\linewidth]{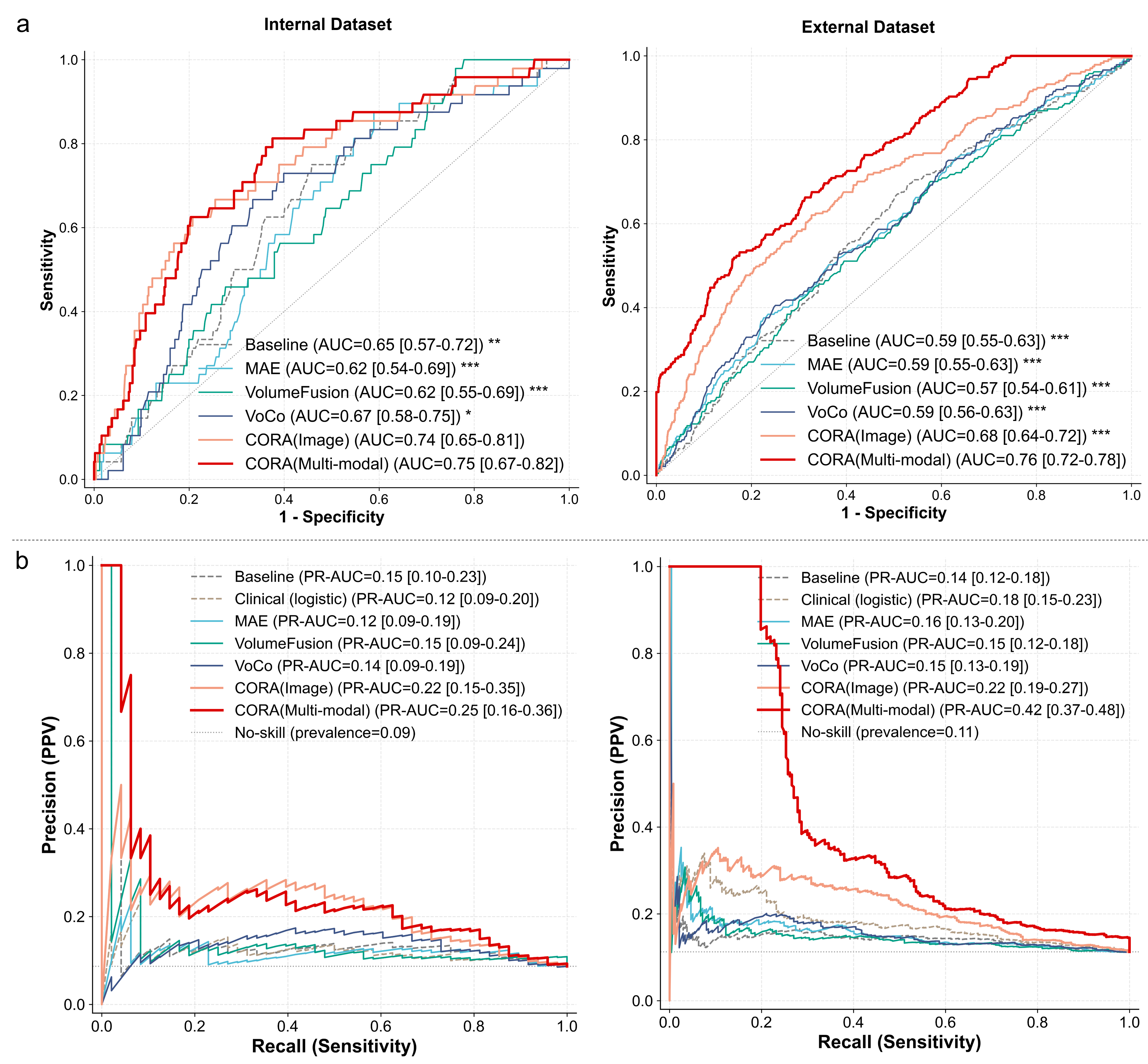}
    }
    \caption{Near-term MACE risk stratification on internal and external test cohorts. (a) Receiver-operating-characteristic curves  and (b) precision–recall curves for 30-day MACE prediction across internal and external cohorts. Compared models include a clinical-variables-only logistic regression, image-only models (self-supervised baselines and CORA(Image)), and CORA(Multimodal), which fuses imaging representations with clinical variables. CORA(Image) was consistently more discriminative than the clinical-only model and all baselines; multimodal fusion provided the largest improvement on the external cohort, particularly in precision–recall performance. The no-information level for precision–recall equals the event prevalence (dotted line). Shaded values are 95\% confidence intervals from 1,000 bootstrap resamples. AUROC and PR-AUC were computed from original model scores.
}
    \label{fig-5}
\end{figure*}

We compared three classes of predictor: a clinical-variables-only model (logistic regression over the demographic and clinical features in Table 1), image-only models (CORA and the self-supervised baselines, each followed by a linear head), and a multimodal variant of CORA that integrates CCTA-derived imaging representations with the same clinical variables encoded through a frozen Qwen-7B language model. Because MACE is a low-prevalence outcome, we report both discrimination (AUROC) and the area under the precision–recall curve (PR-AUC, with the no-information level equal to event prevalence), with 95\% confidence intervals from 1,000 bootstrap resamples (Fig.~\ref{fig-5}).

Across both cohorts, the image-only CORA representation was consistently more discriminative than the clinical-only model and than all self-supervised baselines. On the internal test set, CORA(Image) reached an AUROC of 0.74 (95\% CI: 0.66–0.81), exceeding the clinical-only model (0.61, 0.54–0.69) and the strongest baseline VoCo (0.67, 0.59–0.74). On the external test set, CORA(Image) achieved an AUROC of 0.68 (0.64–0.72) versus 0.64 (0.60–0.68) for the clinical-only model, with all self-supervised baselines at or below 0.59. These results indicate that the learned imaging representation captures prognostic signal beyond that available from structured clinical variables alone.

The added value of multimodal fusion was most evident on the external cohort. There, combining imaging and clinical information raised the AUROC to 0.76 (0.72–0.78) and, more strikingly under class imbalance, increased the PR-AUC from 0.22 (0.19–0.27) for CORA(Image) and 0.18 (0.15–0.23) for the clinical-only model to 0.42 (0.37–0.48) for CORA(Multimodal) — nearly doubling precision–recall performance relative to either single modality. On the internal cohort, by contrast, fusion yielded only a marginal change (AUROC 0.74 to 0.75; PR-AUC 0.22 to 0.25), indicating that in this relatively homogeneous single-site test set with few events (~48 MACE cases) the imaging representation already accounted for most of the available prognostic signal.

The disproportionately larger multimodal gain on the external cohort, which had a higher event rate (11.2\% vs. 8.6\%) and a more heterogeneous demographic composition, suggests that structured clinical variables contribute most when applied to populations that differ from the single-site training distribution. Together, these findings indicate that CORA's imaging representations are prognostically informative and complementary to standard clinical risk factors for near-term MACE, while also delineating where each modality contributes.

\section*{Discussion}
In this study, we introduced CORA, a synthesis-driven 3D vision foundation model designed for comprehensive coronary artery disease assessment and risk stratification from CCTA. By leveraging a pathology-centric self-supervised learning paradigm and large-scale unlabeled data, CORA demonstrated consistent performance gains across a diverse spectrum of downstream tasks, including plaque type characterization, lesion-level stenosis detection, coronary artery segmentation, and short-term MACE prediction. Notably, these improvements were observed not only in internal evaluations but also in independent external cohorts, underscoring the robustness and generalizability of the learned representations. Collectively, our findings suggest that pathology-aware pretraining can serve as a scalable alternative to annotation-intensive supervised learning for cardiovascular imaging.

A central contribution of this work lies in reframing representation learning from generic anatomical reconstruction toward pathology-oriented understanding. Conventional self-supervised paradigms, such as masked image modeling, implicitly emphasize global anatomical statistics and reconstruction fidelity. While effective for structural tasks, these objectives may dilute the representation of spatially sparse pathological signals, particularly in CCTA where coronary plaques occupy only a small fraction of the volumetric field of view. In contrast, the synthesis-driven objective adopted in CORA explicitly biases the encoder toward detecting simulated vascular abnormalities. This formulation encourages the model to allocate representational capacity to subtle, localized intensity and morphological variations that characterize plaque phenotypes. Furthermore, the multi-windowing input strategy enables simultaneous exploitation of complementary attenuation ranges, facilitating improved discrimination between low-attenuation non-calcified plaques and high-density calcified lesions. The consistent gains observed across both diagnostic and anatomical tasks suggest that pathology-centric objectives can simultaneously enhance disease sensitivity and structural awareness.

Compared with existing 3D self-supervised pretraining models and task-specific baselines, CORA demonstrated superior cross-domain generalization, with the largest performance margins observed on external datasets. This finding highlights an important limitation of conventional self-supervised approaches in medical imaging: features learned through generic reconstruction objectives may remain entangled with scanner- or protocol-specific characteristics, limiting transferability across institutions. By contrast, lesion synthesis introduces task-relevant invariances, enabling the model to focus on physiologically meaningful patterns that are less sensitive to acquisition variability. Additionally, the unified encoder architecture supports multiple downstream tasks without reliance on complex multi-stage pipelines, potentially reducing cumulative error propagation associated with vessel extraction and reformatting workflows.

It is important to distinguish the role of synthetic lesions in CORA from conventional synthetic data augmentation. In standard augmentation, synthetic samples are mixed into a labeled training set to enlarge its effective size, and the supervisory signal still derives from human annotations. In CORA, by contrast, the inserted lesions are not auxiliary training examples but constitute the self-supervised objective itself: the location of each synthetic abnormality serves as the sole supervisory target during pretraining, in the complete absence of any diagnostic label. The pretraining task is therefore to localize inserted abnormalities, which directly shapes the geometry of the learned representation space toward sparse vascular pathology. This distinction explains why the observed improvements reflect the pretraining objective rather than mere exposure to synthetic appearances: all compared methods were pretrained on the identical unlabeled cohort under matched conditions, differing only in their self-supervised objective.

\begin{figure*}[h!]
    \centering{  
    \includegraphics[width=1\linewidth]{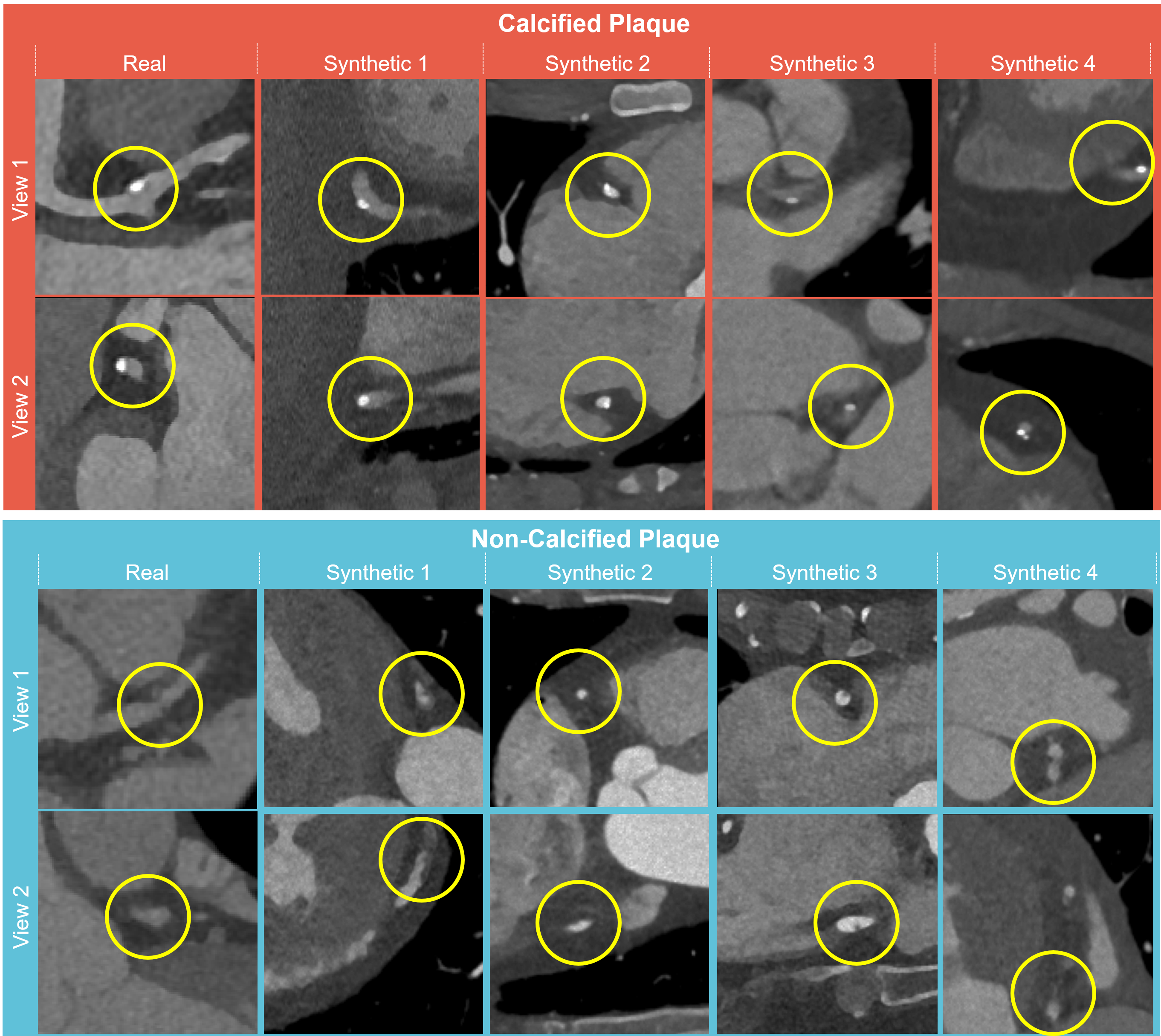}
    }
    \caption{Comparative visualization of real and synthetic coronary plaques. Representative CCTA patches showing calcified and non-calcified  plaques from the clinical dataset. Corresponding examples of synthetic calcified and non-calcified lesions generated by the CORA pathology-aware data engine. The synthetic plaques exhibit a diverse range of morphological structures and Hounsfield Unit (HU) distributions, closely mimicking the attenuation profiles and irregular geometries observed in the real-world clinical cohort.
}
    \label{fig-synthetic data}
\end{figure*}

\begin{figure*}[h!]
    \centering{  
    \includegraphics[width=1\linewidth]{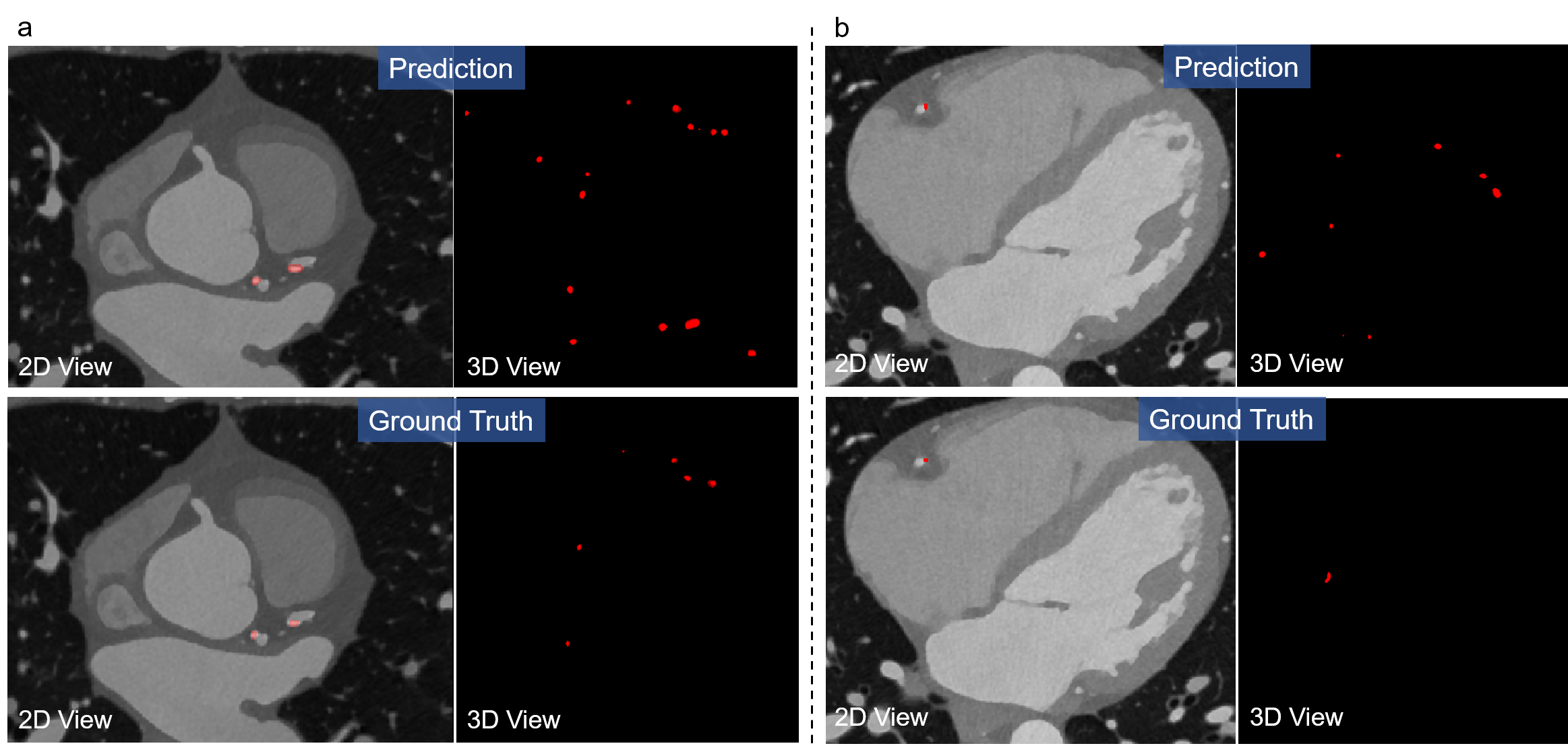}
    }
    \caption{Anatomical attention prior established by synthesis-driven pretraining. Direct inference on real CCTA volumes with (a) calcified and (b) non-calcified plaques using the pretrained CORA model, prior to any task-specific fine-tuning. Raw abnormality predictions (top) are compared to expert-annotated ground truth plaques (bottom) in both 2D axial and 3D spatial views. The pretrained model successfully highlights a broad set of suspected candidate regions (red). Crucially, while this raw output includes false positives, the true clinical lesions are consistently localized within these candidate sets. This demonstrates how pretraining sensitizes the encoder to "lesion-like" features, establishing a highly sensitive prior that significantly narrows the search space for downstream fine-tuning. 
}
    \label{fig-zero}
\end{figure*}

An important consideration concerns the fidelity of synthetic lesions relative to the full biological complexity of atherosclerotic plaques. Real-world coronary lesions exhibit a rich spectrum of morphological and compositional heterogeneity that cannot be exhaustively captured by any simulation engine, including lipid-rich necrotic cores, fibrous caps of variable thickness, spotty calcification, and napkin-ring signs.  
However, we emphasize that perfect pathological realism is neither the design goal nor a prerequisite for effective pretraining. The synthesis engine serves a specific representational purpose: to redirect the encoder's attention from dominant background anatomy toward the coronary vasculature and its potential abnormalities, thereby reducing the learning burden during subsequent task-specific fine-tuning. 
From this perspective, the critical property of synthetic lesions is not biological fidelity but rather morphological and attenuation diversity, ensuring that the pretrained encoder develops broad sensitivity to a wide range of intensity perturbations along the vessel wall. 
A comparative visualization of real and synthetic coronary plaques is provided in Fig.~\ref{fig-synthetic data}.
Empirically, when we applied the pretrained model directly to real CCTA volumes without any fine-tuning, the resulting abnormality response maps highlighted numerous candidate regions, with genuine lesions consistently included among them. Although these raw predictions also contained substantial false positives, this behavior is expected and, in fact, desirable: it indicates that the pretrained encoder has acquired a deliberately over-sensitive detector that casts a wide net over plaque-like regions. Task-specific fine-tuning then refines this broad sensitivity into precise, clinically actionable predictions by leveraging labeled examples to suppress false positives and sharpen lesion discrimination. 
As shown in Fig.~\ref{fig-zero}, we performed zero-shot abnormality detection by applying the pretrained CORA directly to real CCTA volumes without any task-specific fine-tuning.

Furthermore, a notable property of our pretraining framework is its complete independence from diagnostic labels. Because the supervisory signal derives entirely from synthetically inserted lesions, CORA can be pretrained on any CCTA volume regardless of whether the patient harbors genuine coronary atherosclerosis. 
In practice, a subset of the 12,801 pretraining volumes inevitably contains real plaques that are not labeled as positive targets during pretraining. However, several design properties mitigate the impact of this implicit label noise. The patch-based sampling strategy (96$\times$96$\times$96 voxels) ensures that the vast majority of randomly cropped patches do not intersect with real coronary plaques, given their well-established spatial sparsity within the volumetric scan. Furthermore, the large batch size and dataset scale ensure that any noisy gradients from such patches are highly diluted across training iterations. The strong downstream performance across all evaluated tasks provides empirical confirmation that this noise does not meaningfully impair representation quality. More broadly, this property makes the pretraining pipeline immediately scalable to any institutional CCTA archive without curation overhead.

From a clinical perspective, these results have several implications. First, accurate plaque characterization and stenosis detection are critical for diagnostic and therapeutic decision-making decision-making in patients with suspected coronary artery disease. The improved sensitivity and robustness of CORA may facilitate automated screening and prioritization of high-risk studies, thereby reducing interpretation time and inter-observer variability. 
Second, CORA's imaging representations carried prognostic signal for near-term MACE that exceeded a clinical-variables-only model across both cohorts, and integrating the two modalities further improved precision–recall performance, most notably on the external cohort. The larger multimodal gain externally suggests that structured clinical variables are most informative when populations diverge from the training distribution, whereas in a homogeneous single-site setting the imaging representation alone captures most of the available signal. This pattern indicates that imaging-derived and clinical features are complementary rather than redundant, while also tempering claims of uniform multimodal benefit. 
Such integration could support early triage in acute chest pain settings and enable more personalized management strategies. Finally, the unified nature of the proposed foundation model may provide a practical pathway toward scalable deployment across heterogeneous clinical environments.

Several limitations of the current work warrant acknowledgment. First, the clinical composition of our dataset inherently limits the generalizability of the model to asymptomatic populations. In routine clinical practice, CCTAs are primarily acquired for patients with specific clinical indications, most commonly symptomatic chest pain or anginal equivalents. 
Consequently, the generalizability of CORA to asymptomatic screening populations cannot be assumed without dedicated evaluation. The atherosclerotic burden, plaque morphology distribution, and pretest probability in such cohorts may differ substantially from the symptomatic populations represented in our training and validation data. Future prospective studies should assess model performance in asymptomatic individuals undergoing CCTA as part of emerging primary prevention protocols.
Second, our evaluation was based on a retrospective study design, which may not capture the full clinical context of a prospective longitudinal follow-up. Consequently, the labels for 30-day MACE may contain a degree of noise; for instance, out-of-hospital cardiac-related mortality or events occurring at external institutions might not be fully captured within our integrated healthcare system's records. While the probability of such missing events within a narrow 30-day window is relatively low, this potential for under-reporting remains a factor. Future prospective studies with dedicated event adjudication are necessary to further validate these prognostic findings.
Third, our dataset lacked comprehensive baseline data regarding concurrent lipid-lowering therapies (e.g., statin medications). Statins are known to modify plaque biology, often promoting macroscopic calcification while reducing non-calcified lipid cores to stabilize plaques. While the absence of this pharmacological context is a limitation, the strong prognostic performance of CORA across both internal and external cohorts, regardless of underlying medication status, underscores the robustness and broad generalizability of the learned representations in real-world, highly heterogeneous clinical settings.

In conclusion, CORA demonstrates that pathology-centric synthesis-driven pretraining provides an effective strategy for learning clinically meaningful representations from large-scale unlabeled CCTA data. By unifying anatomical assessment and prognostic modeling within a single foundation framework, this approach offers a scalable pathway toward comprehensive, data-efficient cardiovascular imaging analysis.

\bibliography{sample}

\section*{Methods}

\subsection*{Study population}
We retrospectively analyzed a cohort of 12,801 patients who underwent CCTA between October 2005 and November 2024 across nine hospitals within the Northwestern Medicine healthcare system. The study was approved by the Institutional Review Board of Northwestern University, with a waiver for informed consent. Clinical indications for CCTA included symptomatic chest pain, risk-factor-based screening for coronary artery disease (CAD), electrocardiographic abnormalities, or evaluation of prior revascularization.
CCTA examinations were performed using a 64-detector row or higher single- or dual-source CT scanner (Somatom Force or Somatom Definition Flash, Siemens Healthineers; Revolution CT or Revolution Apex, GE HealthCare; Aquilion Prime 80, Canon Medical Systems). All CCTA volumes were identified for analysis. 
All the CCTA volume in pretraining and fine-tuning phase were resampled to a fixed voxel spacing of 0.5 $\times$ 0.5 $\times$ 0.5 $mm^3$. 

To establish the dataset for plaque characterization and MACE prediction, we applied the following exclusion criteria:
1) History of coronary revascularization.
2) Coronary artery bypass graft surgery.
3) Prior acute myocardial infarction.
4) Absence of follow-up diagnostic records following the index CCTA.
The final study population consisted of 4,883 eligible subjects with an average follow-up of 2.4 years. Of these, 2,775 patients were from Northwestern Memorial Hospital, partitioned into a training set ($N=2,220$) and an internal validation set ($N=555$). The remaining 2,108 patients from the other eight hospitals were reserved as an independent external validation cohort.

To strictly prevent data leakage, all patients constituting the internal test set (N=555) and the external test set (N=2,108) were excluded from the pretraining set prior to self-supervised pretraining, yielding 10,138 pretraining volumes. No test-set image was observed by the model during pretraining; the stenosis detection cohort (N=348) and the ImageCAS test split were likewise independent of the pretraining data. Only the downstream training partition overlaps with the pretraining cohort, consistent with standard self-supervised pretraining followed by supervised fine-tuning.

\begin{figure*}[h!]
    \centering{  
    \includegraphics[width=0.9\linewidth]{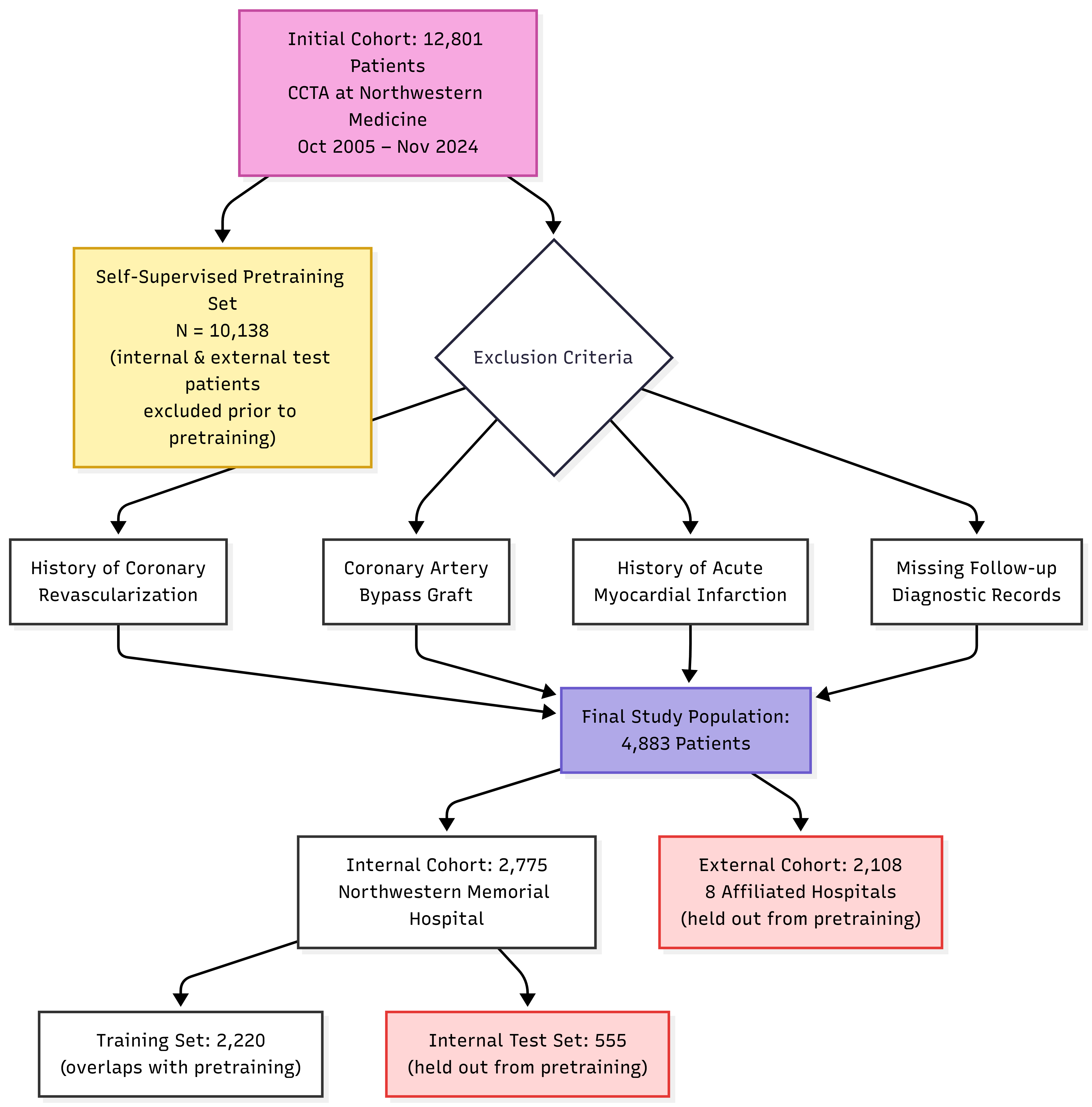}
    }
    \caption{Flowchart of study population selection and cohort partitioning. From an initial cohort of 12,801 patients undergoing CCTA at Northwestern Medicine, the self-supervised pretraining set (N = 10,138) was constructed after excluding all internal and external test patients, ensuring that no test-set volume was seen during pretraining. For downstream evaluation, 4,883 patients were retained after applying clinical exclusion criteria, then partitioned into an internal cohort from Northwestern Memorial Hospital (N = 2,775; 2,220 training, 555 internal test) and an independent external cohort (N = 2,108) from eight affiliated hospitals. Only the training partition overlaps with the pretraining set; both test cohorts were held out.}
    \label{fig-data}
\end{figure*}

The demographic, clinical, and imaging characteristics for each cohort are summarized in Table~\ref{tab-1}. The detailed inclusion and exclusion process, including patient partitioning across the nine participating centers, is illustrated in the study flow diagram (Fig.~\ref{fig-data}).

Plaque classification labels were extracted from the original free-text CCTA radiology reports using two large language models independently, GPT-4o and Claude Sonnet 4.5, each prompted to return structured JSON indicating the presence of calcified and non-calcified plaque. The two model outputs were compared, and any discordant cases were resolved by manual expert review of the corresponding report. On a randomly sampled subset of 200 reports adjudicated by a board-certified reader, both models achieved extraction accuracy exceeding 99\%. The extraction prompts and output schema are available in the code repository (see Code Availability)

The stenosis detection cohort comprised 348 CCTA volumes drawn from the Northwestern Medicine system, none of which were included in the pretraining set. We identified 348 CCTA volumes that contained at least one lesion with mild or greater severity (CAD-RADS $\ge$ 2), as indicated in the original clinical radiology reports. To ensure diagnostic-grade ground truth, three experienced radiologists and medical imaging experts independently annotated the stenotic lesions, using the corresponding radiology reports as a primary reference for localization and grading.

\subsection*{CORA implementation and pretraining protocol}
We framed the pretraining of CORA as a self-supervised segmentation task, utilizing a 3D Residual U-Net architecture (Table~\ref{tab-architecture}). To prioritize the learning of clinically relevant features, we developed a "pathology-aware" data engine that performs real-the-fly synthesis of coronary abnormalities within the vascular tree. This strategy biases the model’s representational capacity toward the coronary vasculature and its associated pathologies, rather than generic anatomical reconstruction.

\begin{table}[ht!]
\centering
\caption{Architectural specifications of the CORA model during the pretraining. The network follows a 3D Residual U-Net design with a four-stage encoder-decoder structure.}
\label{tab-architecture}
\small
\begin{tabular}{llcccc}
\toprule
\textbf{Part} & \textbf{Stage} & \textbf{Operation} & \textbf{Output Features} & \textbf{Stride} & \textbf{Blocks} \\
\midrule
\textbf{Input} & - & CCTA Multi-channel Patch & $4 \times 96^3$ & - & - \\
\midrule
\textbf{Encoder} & Stage 1 & 3D ResBlock & $32 \times 96^3$ & $[1, 1, 1]$ & 1 \\
& Stage 2 & 3D ResBlock & $64 \times 48^3$ & $[2, 2, 2]$ & 3 \\
& Stage 3 & 3D ResBlock & $128 \times 24^3$ & $[2, 2, 2]$ & 4 \\
& Stage 4 & 3D ResBlock (Bottleneck) & $256 \times 12^3$ & $[2, 2, 2]$ & 4 \\
\midrule
\textbf{Decoder} & Stage 3 & Upsampling + 3D Conv & $128 \times 24^3$ & $[2, 2, 2]$ & 1 \\
& Stage 2 & Upsampling + 3D Conv & $64 \times 48^3$ & $[2, 2, 2]$ & 1 \\
& Stage 1 & Upsampling + 3D Conv & $32 \times 96^3$ & $[2, 2, 2]$ & 1 \\
\midrule
\textbf{Output} & - & $1 \times 1 \times 1$ Convolution & $1 \times 96^3$ & - & Sigmoid \\
\bottomrule
\addlinespace
\multicolumn{6}{l}{\footnotesize \textit{Note}: All convolutional layers use $3 \times 3 \times 3$ kernels followed by Instance Normalization and LeakyReLU.} \\
\end{tabular}
\end{table}

\textbf{Data engine} To facilitate the data engine, we first extracted coronary artery masks from the pretraining volumes using a nn-unet~\cite{isensee2021nnu} model trained on the ImageCAS dataset~\cite{zeng2023imagecas}.
During the pretraining, a artery-centric sampling strategy: patches of size $96 \times 96 \times 96$ were randomly cropped, with the center of each patch anchored to a coronary artery voxel. This ensures that every training sample contains valid vascular anatomy, maximizing the learning signal for coronary-specific features.

Within each vascular patch, we stochastically inserted synthetic calcified or non-calcified (soft) plaques. We utilized a data engine to generate irregular, realistic lesion morphologies. Lesions were modeled as composite structures formed by 1–3 overlapping Gaussian blobs (sigma range: $0.7-2.0$) to simulate the heterogeneous shapes of natural plaques.
Synthetic lesions were assigned Hounsfield Unit (HU) values according to their clinical phenotype: $800-1500$ HU for calcified plaques and $30-90$ HU for soft plaques.
To provide the model with enhanced contrast sensitivity, the raw patch was transformed into a four-channel input using specific clinical windows:
Fat Window ($-100, 140$ HU), Soft tissue Window ($50, 400$ HU), Angiographic Window ($350, 700$ HU) and Calcification Window ($500, 2000$ HU).
To enhance robustness and simulate clinical imaging conditions, we applied Poisson noise using a Beer-Lambert law approximation ($I_0 = 10^5, L = 200.0$ mm), alongside electronic background noise ($\sigma_e = 2.0$). Geometric augmentations included random 3D rotations ($\pm 15^\circ$), isotropic zooming ($0.9-1.1$), and random axis flipping.

\textbf{Model pretraining} Given that coronary lesions are highly sparse relative to the total volume, we utilized a custom loss function to address extreme class imbalance. The objective function is defined as the sum of two components: 1) Tversky Loss\cite{salehi2017tversky}: We set $\beta = 0.9$ and $\alpha = 0.1$ to heavily penalize false negatives, thereby prioritizing recall (sensitivity) in detecting subtle lesions.
2) Focal Loss~\cite{lin2017focal}: A focusing parameter of $\gamma = 4.0$ was employed to down-weight the loss contribution from easy background voxels, forcing the model to concentrate on the challenging pathological boundaries.
The total loss is:
$$L_{total} = L_{Tversky}(\alpha=0.1, \beta=0.9) + L_{Focal}(\gamma=4.0)$$
This synthesis-driven pretraining biases the encoder toward clinically relevant vascular pathology, providing a robust foundation for downstream diagnostic and prognostic tasks.

\textbf{Baseline pretraining protocol} To ensure a fair and controlled comparison, all self-supervised baselines (MAE, VolumeFusion, and VoCo) were pretrained from scratch on the identical cohort of 10,138 unlabeled CCTA volumes used for CORA, rather than initialized from publicly available checkpoints pretrained on external data. All baselines shared the same backbone capacity, input patch size and voxel spacing, and were trained under matched computational budgets on the same hardware. The only variable that differed across methods was the self-supervised objective itself: masked-reconstruction-based learning for the baselines versus synthesis-driven abnormality detection for CORA. This design isolates the effect of the pretraining objective from confounds related to data scale, model capacity, input representation, or compute.
The training-from-scratch baseline used the identical 3D Residual U-Net architecture as CORA with randomly initialized weights, and was trained under the same downstream fine-tuning configuration as all other methods; it differed from the pretrained models solely in the absence of any self-supervised pretraining.

\subsection*{Adaptation to downstream tasks}
\begin{figure*}[ht!]
    \centering{  
    \includegraphics[width=1\linewidth]{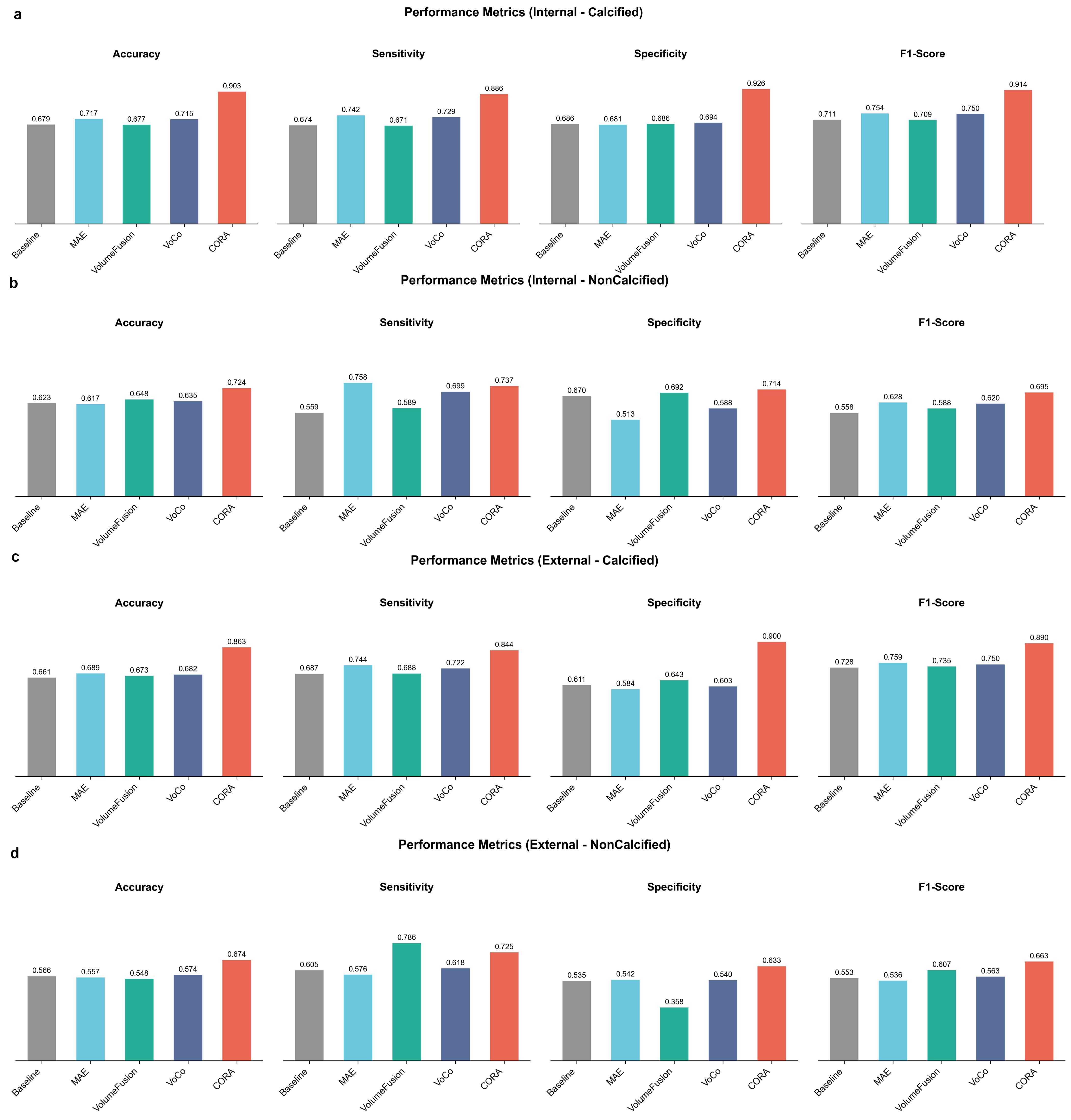}
    }
    \caption{Detailed quantitative results of coronary plaque characterization task.}
    \label{fig-sup-plaque}
\end{figure*}

\textbf{Coronary plaque characterization task}
The coronary plaque characterization task was formulated as a volume-level multi-label classification problem to predict the presence of calcified and non-calcified plaques from CCTA volumes.
During fine-tuning, we used the pretrained encoder of CORA as the backbone, followed by an global average pooling layer and a linear classification head. As this task operates at the volume level, the entire CCTA volume was used as input without patch sampling.
Model performance was primarily evaluated using the area under the receiver operating characteristic curve (AUROC), as reported in the main text. In addition, we reported accuracy, sensitivity, specificity, and F1-score for both calcified and non-calcified plaque detection (Fig.~\ref{fig-sup-plaque}). 

\textbf{Stenosis detection}
The stenosis detection task was formulated as a segmentation problem.
For fine-tuning, both the pretrained encoder and decoder of CORA were used as initialization. To address severe class imbalance between stenotic lesions and background, we adopted a composite loss function consisting of Tversky loss and focal loss.
To reduce false-positive detections and improve computational efficiency during the fine-tuning stage, we utilized TotalSegmentator~\cite{wasserthal2023totalsegmentator} to generate an anatomical mask of the heart. This mask was then used to crop the original CCTA volumes, effectively isolating the cardiac region and restricting the model’s input to clinically relevant voxels. This preprocessing step significantly suppressed potential artifacts from extracardiac structures.
Because stenosis boundaries, especially those caused by non-calcified plaques, are often ambiguous, and clinical decision-making primarily requires lesion localization rather than precise voxel-level delineation, evaluation was conducted at the lesion level. Each stenosis was defined as a connected component in the ground-truth annotation. A predicted lesion was considered a true positive if it overlapped with a reference lesion by more than 10 voxels.

\textbf{Coronary artery segmentation}
To evaluate the transferability of our pre-trained representations to dense anatomical tasks, we fine-tuned CORA for coronary artery segmentation. In this configuration, we utilized the pre-trained CORA encoder paired with a randomly initialized decoder. The model was optimized using Dice loss to ensure overlap maximization between the predicted and reference vascular masks. This task specifically tested whether pathology-centric pretraining preserves the underlying anatomical grounding of the coronary tree.

\textbf{MACE risk stratification}
For 30-day MACE prediction, we constructed a multimodal model integrating imaging and clinical information. The pretrained CORA encoder, followed by global average pooling, was used to obtain a fixed-length imaging representation. In parallel, the demographic and clinical variables listed in Table~\ref{tab-1} were serialized into a natural-language clinical summary on a per-patient basis. Rather than presenting variables as isolated field–value pairs, each patient's structured fields were rendered into fluent English sentences organized into five thematic segments: a task-orienting prompt, demographics (age, sex), vital signs (blood pressure, body mass index), laboratory values (LDL, HDL, total cholesterol), medical history (diabetes, hypertension; included only when present), and smoking status (mapped from never/former/current to descriptive phrasing). A representative serialized input is: \textit{The following is a summary of the patient's demographic and clinical information. Please focus on key risk factors for cardiovascular disease. This is a 63-year-old male. The patient's blood pressure is 128/73 mmHg, BMI is 29.0 kg/m\textsuperscript{2}. Laboratory results show LDL of 96 mg/dL, HDL of 50 mg/dL, total cholesterol of 170 mg/dL. Medical history includes diabetes, hypertension diagnosed within the past year. Regarding lifestyle, the patient is a former smoker.''} Missing numerical fields were filled with training-set mean values to maintain consistency with the imputation used for the clinical-only baseline, and categorical fields without a recorded value were rendered with an explicit unknown'' descriptor; history items absent for a given patient were simply omitted from the summary.

The serialized text, with its task-orienting prompt prepended as the leading sentence, was encoded with a frozen Qwen-7B language model. The pooled text embedding was projected through a multilayer perceptron (MLP) to match the imaging feature dimension, concatenated with the imaging representation, and passed to a linear classification layer. The model was trained with cross-entropy loss; only the MLP, fusion, and classification layers were updated, while both the imaging encoder and the language model remained frozen.

We adopted a language-model text encoder, rather than a conventional tabular encoder, so that heterogeneous structured fields and missing values could be represented within a single unified embedding space through natural-language serialization, and so that the clinical encoder could be extended to richer free-text inputs (e.g., clinical notes) in future work without architectural change. We note that the marginal contribution of this clinical encoder over imaging alone varied by cohort (Results), which we examine further in the Discussion.

\subsection*{Computational resources}
All model training and evaluations were performed on the Northwestern University Quest High-Performance Computing (HPC) cluster. The pretraining phase of CORA was conducted using a single NVIDIA H100 GPU ($80$ GB). Due to the efficient patch-based sampling strategy ($96 \times 96 \times 96$ voxels), the pretraining process was completed in approximately $25$ hours. For downstream fine-tuning tasks, the computational requirements varied based on input dimensionality and dataset scale. 
Plaque Classification and MACE Risk Stratification: Because these tasks required processing the entire CCTA volume as input, we utilized a multi-GPU configuration consisting of three NVIDIA H100 GPUs. Training for these tasks was completed in approximately $48$ hours. 
Coronary Artery Segmentation and Stenosis Detection: These tasks were computationally less intensive due to the smaller dataset sizes. Fine-tuning for these objectives was performed on a single NVIDIA H100 GPU, with each task reaching convergence in less than $10$ hours.
All implementations were developed using the PyTorch framework, leveraging specialized libraries such as SimpleITK and MONAI for medical image processing and data augmentation.

\section*{Data availability}

The ImageCAS dataset was used for the evaluation of coronary artery segmentation and is publicly available at 
\href{https://github.com/XiaoweiXu/ImageCAS-A-Large-Scale-Dataset-and-Benchmark-for-Coronary-Artery-Segmentation-based-on-CT}{GitHub repository}.
The datasets used for pretraining CORA, as well as for fine-tuning on the coronary plaque characterization task, stenosis detection task, and MACE risk stratification task, are not publicly available due to patient privacy obligations, Institutional Review Board (IRB) restrictions, and data use agreement requirements.
De-identified data may be made available for research purposes from the corresponding authors upon reasonable request, subject to institutional approval.

\section*{Code availability} 
All algorithms used in this study were developed using libraries and scripts in PyTorch. The source code will be publicly available at: 
\href{https://github.com/Advanced-AI-in-Medicine-and-Physics-Lab/CORA.git}{[CORA]}.

\section*{Author contributions statement}
J.H. proposed the methodology, designed and performed all experiments, conducted the data analysis, and drafted the manuscript. G.D. and H.E.A. performed the data annotation and verified the integrity of the clinical datasets. B.Z. conceptualized, managed data resources, and supervised the research project. U.B., B.D.A., and N.S.S. provided critical clinical and technical insights. All authors contributed substantially to the critical revision of the manuscript for important intellectual content, and all authors reviewed and approved the final version.

\section*{Competing interests} 
The authors declare no competing interests. 





\end{document}